%% file: main.tex
\documentclass[letterpaper, 10 pt, conference]{ieeeconf}  

\IEEEoverridecommandlockouts                              

\usepackage{graphics} 
\usepackage{epsfig} 
\usepackage{amsmath} 
\usepackage{amssymb}  
\usepackage{booktabs}       
\usepackage{graphicx}
\usepackage{textcomp}
\usepackage{xcolor}
\usepackage{bbm}
\usepackage{wrapfig}
\usepackage{cite}
\usepackage{algorithm}
\usepackage{algorithmic}
\usepackage{dsfont}
\usepackage[hidelinks]{hyperref}

\usepackage{subcaption}
\usepackage{amsthm}

\usepackage[capitalize]{cleveref}
\crefname{section}{Sec.}{Secs.}
\Crefname{section}{Section}{Sections}
\Crefname{table}{Table}{Tables}
\crefname{table}{Tab.}{Tabs.}
\DeclareFontFamily{OT1}{cmtt}{\hyphenchar \font=-1}

\newcommand\numberthis{\addtocounter{equation}{1}\tag{\theequation}}
\newcommand{\link}[1]{\textcolor{magenta}{\href{#1}{#1}}}

\newcommand{\M}{\mathcal{M}}
\renewcommand{\S}{\mathcal{S}}
\newcommand{\A}{\mathcal{A}}

\renewcommand{\P}{\mathcal{P}}

\newcommand{\D}{\mathcal{D}}
\newtheorem{assumption}{Assumption}
\newcommand{\Method}{\texttt{Rank2Reward\,}}

\title{\LARGE \bf Rank2Reward: Learning Shaped Reward Functions from Passive Video}

\author{Daniel Yang$^{1}$, Davin Tjia$^{2}$, Jacob Berg$^{2}$, Dima Damen$^{3}$, Pulkit Agrawal$^{1}$, Abhishek Gupta$^{2}$
\thanks{$^{1}$D. Yang and P. Agrawal are with the Computer Science and Artificial Intelligence Laboratory, Massachusetts Institute of Technology, Cambridge, MA, USA \texttt{\{dxyang, pulkitag\}@mit.edu}}%
\thanks{$^{2}$D. Tjia, J. Berg, and A. Gupta are with the Department of Computer Science, University of Washington, Seattle, WA, USA \texttt{\{davin05, jacob33, abhgupta\}@cs.washington.edu}}%
\thanks{$^{3}$D. Damen is with the School of Computer Science, University of Bristol, Bristol, UK \texttt{dima.damen@bristol.ac.uk}}%
}

\begin{document}

\maketitle
\thispagestyle{empty}
\pagestyle{empty}

\begin{abstract}

Teaching robots novel skills with demonstrations via human-in-the-loop data collection techniques like kinesthetic teaching or teleoperation puts a heavy burden on human supervisors. In contrast to this paradigm, it is often significantly easier to provide raw, action-free visual data of tasks being performed. Moreover, this data can even be mined from video datasets or the web. Ideally, this data can serve to guide robot learning for new tasks in novel environments, informing both ``what" to do and ``how" to do it. A powerful way to encode both the ``what" and the ``how" is to infer a well-shaped reward function for reinforcement learning. The challenge is determining how to ground visual demonstration inputs into a well-shaped and informative reward function. We propose a technique \Method for learning behaviors from videos of tasks being performed without access to any low-level states and actions. We do so by leveraging the videos to learn a reward function that measures incremental ``progress" through a task by learning how to temporally rank the video frames in a demonstration. By inferring an appropriate ranking, the reward function is able to guide reinforcement learning by indicating when task progress is being made. This ranking function can be integrated into an adversarial imitation learning scheme resulting in an algorithm that can learn behaviors without exploiting the learned reward function. We demonstrate the effectiveness of \Method at learning behaviors from raw video on a number of tabletop manipulation tasks in both simulations and on a real-world robotic arm. We also demonstrate how \Method can be easily extended to be applicable to web-scale video datasets.
Code and videos are available at \link{https://rank2reward.github.io}
\end{abstract}

\input{sections/intro}
\input{sections/related_works}
\input{sections/problem}
\input{sections/methods}

\input{sections/results}
\input{sections/discussion}

\section{Acknowledgements}
\label{sec:acks}
This research was supported by NSF Robust Intelligence Grant 2212310, the Hyundai Motor Company, and an Amazon Research Award. We thank members of the WEIRD lab at UW and the Improbable AI lab at MIT for helpful discussions and feedback and the HYAK compute cluster as well as MIT Supercloud for compute support. 


\bibliographystyle{IEEEtran}
\bibliography{main}


\addtolength{\textheight}{-12cm}   


\end{document}

%% file: sections/intro.tex
\section{Introduction}
\label{sec:intro}

Robot learning via reinforcement learning (RL) directly in the real world show the promise of continual improvement, with minimal modeling assumptions \cite{kalashnikov2018qt, andrychowicz2020learning, gu2017deep, margolisyang2022rapid, levine2016end}. However, the promise of plug-and-play reinforcement learning hides a significant challenge --- \emph{where do reward functions come from?} Reward function design is a non-trivial task; rewards must be unbiased while still guiding exploration toward optimal behaviors with ``dense supervision". While this may be possible to provide in certain simulation environments \cite{yu2020meta, james2019rlbench}, it is much more challenging in the real world. 

\begin{figure}[ht]
        \centering
        \vspace{-0.3cm}
        \includegraphics[width=0.95\columnwidth]{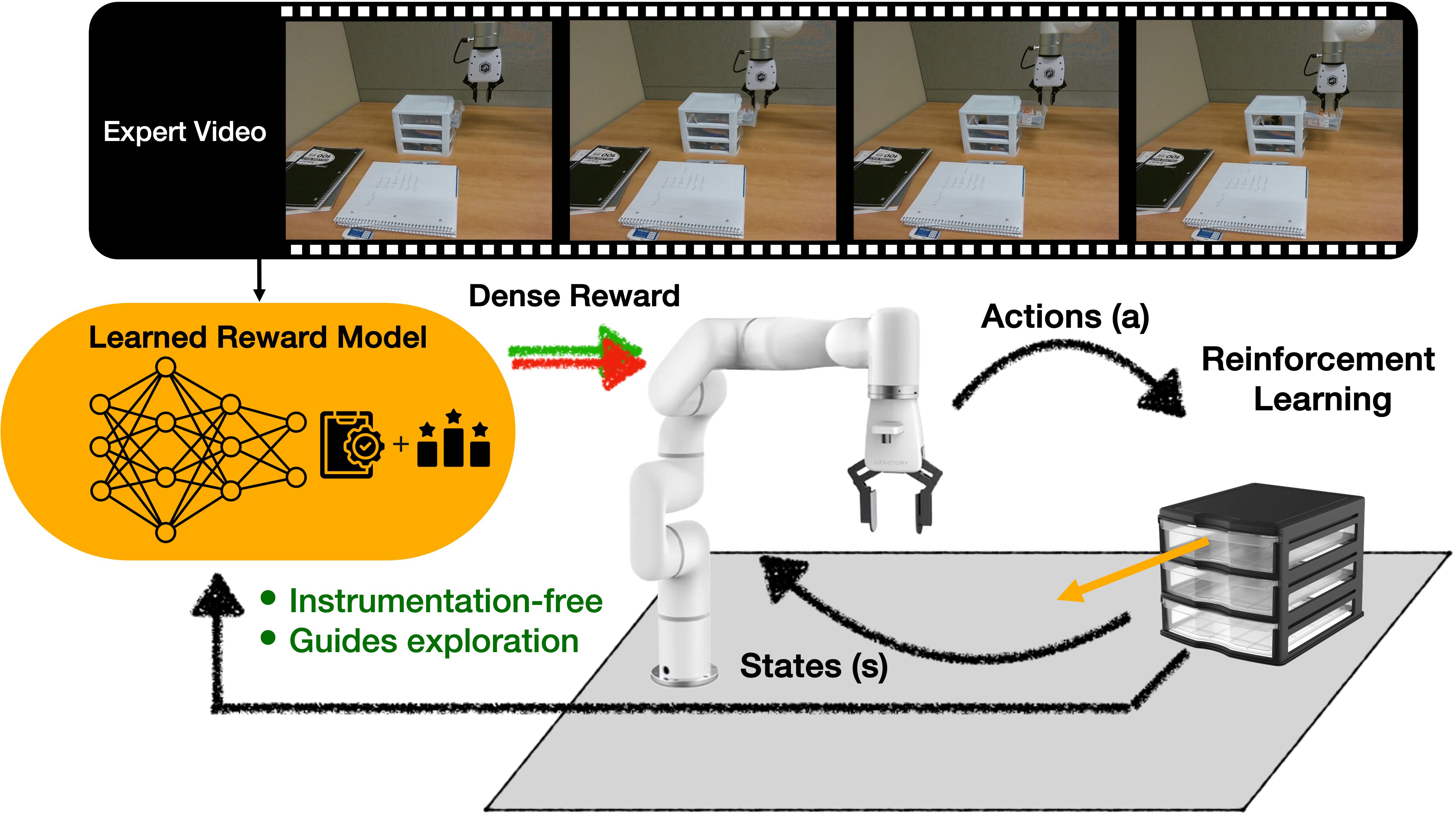}
        \caption{\footnotesize{Depiction of the problem setting in \Method - inferring well-shaped and calibrated reward functions from video demonstrations that enable effective policy optimization.}}
        \label{fig:teaser}
    \vspace{-0.75cm}
\end{figure}

A natural strategy for reward function design is data-driven algorithms such as inverse RL\cite{russell1998learning, ng2000algorithms} for reward inference. These methods rely on expert demonstrations to infer reward functions, learning reward functions that maximize the likelihood of demonstrations while being uninformative about other trajectories \cite{ho_generative_2016, fu_variational_2018, finn16gcl}. This approach can be powerful but typically suffers from two significant challenges - (1) demonstration data in the form of state-action tuples can be challenging to obtain without expensive techniques such as kinesthetic teaching or teleoperation, and (2) learned reward functions may explain the expert data well, but may not be ``well-shaped", providing no guidance for exploration. For data-driven reward functions to be a practical alternative to hand-crafted reward functions, they must both be easy to provide and make policy optimization easy.   

As opposed to expensive forms of demonstrations such as kinesthetic teaching or teleoperation, a natural and easy-to-obtain source of interaction data is video observations of tasks being performed. These are abundantly present in computer vision datasets \cite{Ego4D2022CVPR, Damen2018EPICKITCHENS, goyal2017something, materzynska_something-else_2020}. This data contains both ``what" tasks are interesting in an environment and ``how" to accomplish these tasks. In this work, we show how raw videos of tasks being performed can serve as supervision for a simple reward learning method that satisfies both desiderata above - (1) is easy to provide, and (2) effectively guides exploration for RL by providing informative shaping. 

The key insight that we exploit in this work is that video demonstrations typically make monotonic progress toward a goal. Under this assumption, a natural reward function is simply how much \emph{progress} has been made along a successful trajectory. This framing allows us to recast reward function learning as the problem of learning to order frames within a video. By predicting an ordering of video frames, we can infer a notion of progress along a trajectory using techniques from learning from preferences ~\cite{christiano17pref, ouyang2022instructgpt}. Since the progress along a trajectory is strictly monotonic, the resulting reward function is well-shaped for policy optimization. 

Notably, since the ranking function is trained purely on expert video data, it cannot meaningfully provide a reward signal to states and trajectories not covered in the expert dataset. To remedy this, we show how to formulate policy search with learned ranking rewards as a constrained policy optimization problem in which the policy is constrained to stay close to the distribution of expert data. We show how this can be further simplified to a weighted variant of adversarial imitation learning, alternating between (1) learning a discriminator to differentiate expert from on-policy trajectories and (2) policy search using reward as a combination of the learned ranking function and the learned discriminator. This results in a simple yet performant algorithm for policy learning from video demonstrations without actions - rank frames in video demonstrations and use this ranking to reweight adversarial imitation learning. We show the efficacy of this learned reward function to guide policy learning for both tasks in simulation and real-world robotic manipulation. 

%% file: sections/related_works.tex
\section{Related works}
\label{sec:related_works}
Our work builds closely on a wide variety of related work, as we outline below. 

\textbf{Inverse RL:} Inverse reinforcement learning (IRL), \cite{russell1998learning, ng2000algorithms, abbeel_apprenticeship_2004} aims to infer rewards from demonstrations such that demonstrations are scored highly, while \emph{other} trajectories are scored suboptimally. Various IRL techniques aim to instantiate this idea using techniques such as max-margin planning \cite{ratliff2006maximum}, maximum entropy inverse RL \cite{ziebart2008maxent, fu2017learning, finn16gcl}, and feature matching \cite{abbeel_apprenticeship_2004}. Generative adversarial imitation learning (GAIL) \cite{ho_generative_2016} and similar methods \cite{finn16gcl, fu_variational_2018, singh2019rewardengineering}, proposed treating IRL as an adversarial game. A challenge most IRL techniques face is that while the reward is correct at optimality \cite{cai2019global, chen2020computation}, the rewards are poorly shaped, offering no learning signal. We show that a simple ranking based objective can allow us to easily infer well-shaped rewards.  

\textbf{Imitation from observation:} Imitation-from-observation considers how to learn from high dimensional \emph{action-free} demonstrations such as videos \cite{torabiWS19a}. One class of these imitation-from-observation techniques tries to label the observation-only dataset with inferred actions from an inverse dynamics model, and run standard imitation learning \cite{schmeckpeper_reinforcement_2021, baker2022video, radosavovic_state-only_2021}. Specifically for tasks with human hands, other methods infer actions using off the shelf hand and object pose estimation algorithms before applying standard imitation learning \cite{qin2022dexmv, bahl2022human, shaw2022videodex}, but these methods require instrumentation of the environment with calibrated, depth-sensing cameras and presume known pose detectors. \cite{shao2021concept2robot, chen2021learning} learn video classifiers and use these as rewards for reinforcement learning. This can be effective at deciding ``what" to do but fails to provide shaped rewards. Other techniques to learn from videos include learning representations from video to assign rewards \cite{sermanet2018time}, using temporal contrastive learning \cite{ma23vip, ma2023liv, alakuijala2023learning}, and regressing onto temporal differences between frames to provide an exploration bonus reward for RL \cite{bruce23iclr}. While this can be effective in certain scenarios, these learned distances are inaccurate out of distribution and are prone to exploitation. As opposed to these methods, \Method aims to provide a reward function that is \emph{both} well-shaped and calibrated on out-of-distribution states. The closest work in this line to ours is Time Contrastive Networks (TCN) \cite{sermanet2018time}. While TCN may learn a useful representation from contrastive learning across time and viewpoints, this embedding space does not contain any notion of progress towards achieving a goal, as distance in input space does not correspond to moving towards or away from the goal, and relies on performing feature tracking on a specific expert trajectory which requires temporal alignment. In contrast, \Method learns an ordered ranking space which both encodes progress towards the goal \emph{and} is agnostic to time required to reach the state.

\textbf{Ranking-based approaches in video understanding:} Modeling the evolution of human actions in video through temporal frame ranking was first proposed in~\cite{Fernando_2015_CVPR}. Using a ranking loss, the approach learns a representation of an action that successfully orders the frames in that video. Subsequently, ranking was successfully used for end-to-end classification~\cite{fernando16}, modeling progression to completion~\cite{Federico2020} as well as skill development~\cite{Doughty_2018_CVPR}. In this work, we take inspiration from using the temporal frame ranking loss.

\textbf{Ranking-based reward learning:} Our work leverages a frame ranking objective to infer well-shaped rewards. The idea of ranking based objectives being used has been recently explored in the context of reinforcement learning from human feedback \cite{ouyang2022instructgpt, christiano17pref, biyikpref, dorsapref, brownmbrl}. This can be used to leverage binary comparisons, provided by a human, to train a reward function that can be used for RL. In contrast, our work does not need any external human comparisons or preferences. Instead, we simply rank frames within a video according to their temporal progression.

%% file: sections/problem.tex
\section{Preliminaries}
\label{sec:problem}
Consider a robot learning how to perform a task as a finite horizon Markov decision process (MDP), $\M$, consisting of the tuple $(\S, \A, \P, \rho_0, \gamma)$ where $\S$ is the state space, $\A$ is the action space, $\P(s' | s, a)$ is the transition function, $\rho_0$ is the initial state distribution, and $\gamma$ is the discount factor. In this work, $\S$ can be high-dimensional images, but we assume that the environment is fully observable, thereby retaining the Markov property. This can be relaxed by either stacking frames or leveraging history-conditioned policies \cite{mnih13dqn}.

Here, we want our robot to learn a policy, $\pi^*$, to complete a particular task. However, the true reward function $r(s)$ is not available to the learning agent. Instead, we have a set of $N$ expert demonstration trajectories, $\D^e = \{\tau_k\}_{k=1}^N$ where each trajectory consists of a sequence, $\tau_k = \{s_0^k, s_1^k, \dots, s_T^k \}$. Without loss of generality, we assume that the expert dataset is drawn IID from some expert policy $\pi^e$, with corresponding state-action marginal $d^e(s, a)$. Unlike typical imitation learning settings, no actions are available.

Since the reward function is unknown, the goal is first to infer an appropriate reward function $\hat{r}(s)$ from the expert data and then use this for policy optimization as $\pi^* \leftarrow \arg \max_{\pi} \mathbb{E}_{}\left[ \sum_t \gamma^t \hat{r}(s) \right]$ similar to standard reinforcement learning settings. Using the notation of the state-action marginal, $d^\pi(s, a)$, we can rewrite this policy optimization objective as $\pi^* \leftarrow \arg \max_{\pi} \mathbb{E}_{d^\pi(s, a)}\left[\hat{r}(s) \right]$, where $d^\pi(s, a) = (1 - \gamma) \sum_{t=0}^\infty P(s_t = s, a_t = a| s_0 \sim \rho_0(s), a_t \sim \pi(a_t|s_t), s_{t+1} \sim \P(\cdot| s_t, a_t))$ is the standard state-action occupancy measure. Note that as in most IRL settings, the process of inferring $\hat{r}(s)$ and learning $\pi^*$ can be interleaved. 

%% file: sections/methods.tex
\section{Rank2Reward: Learning shaped reward functions by frame ranking}
\label{sec:method}

\begin{figure*}
    \centering
    \includegraphics[width=0.8\linewidth]{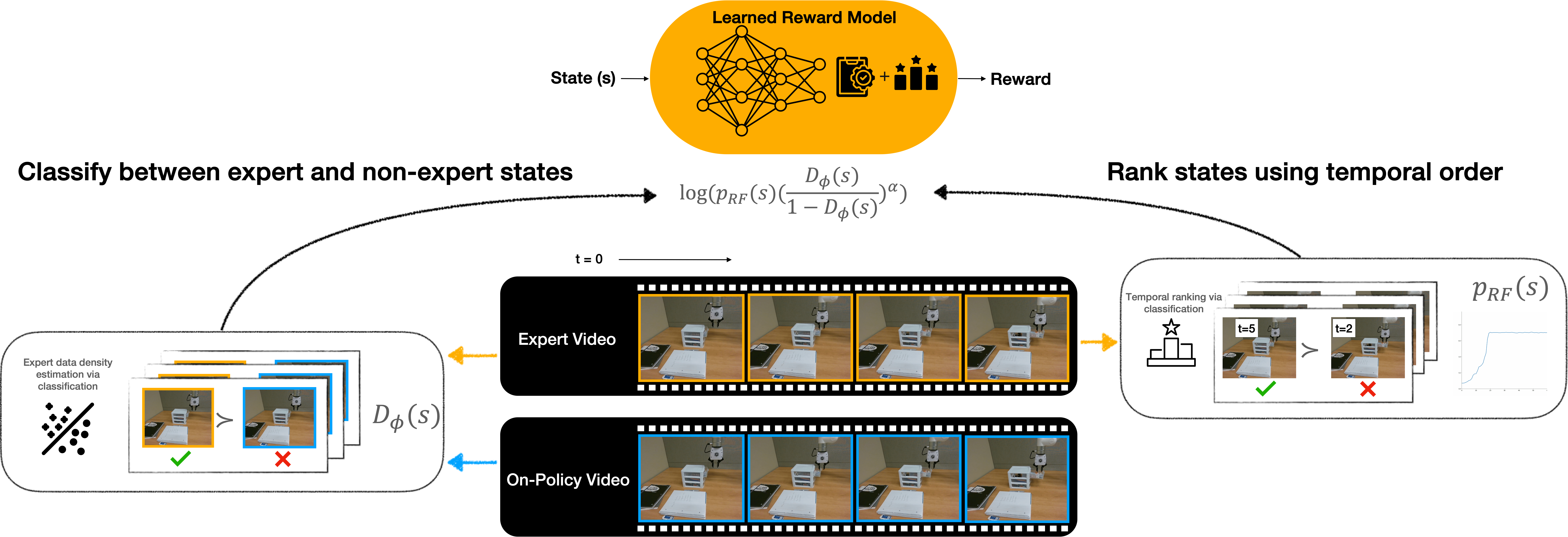}
    \caption{\footnotesize{A schematic depiction of reward inference using \Method. Given video demonstrations from a human supervisor, \Method learns a reward function by combining two distinct elements --- (1) a ranking function that temporally orders frames, providing a monotonically increasing reward signal $p_{RF}(s)$. Secondly, (2) a classifier $D_\phi$ between expert and on-policy data, so that expert data is weighted higher than on-policy data. When combined multiplicateively, they yield a well-shaped reward function for RL that pushes down on-policy data, and pushes up expert data.}}
    \label{fig:mainfig}
    \vspace{-0.5cm}
\end{figure*}

We propose a simple and scalable method for reward function inference from raw video demonstrations without requiring any actions. Our proposed technique, \Method as shown in \cref{fig:mainfig}, can learn \emph{well-shaped} reward functions that guide exploration for challenging tasks while being resilient to exploitation by the policy during reinforcement learning. The key idea is to learn how to order frames temporally in a trajectory. In doing so, we can infer whether states visited by a policy are making \emph{progress} along a trajectory, therefore learning policies that maximize progress. We then show how this objective in itself is prone to exploitation during policy learning and propose a constrained policy learning objective that prevents this exploitation. The resulting algorithm resembles a weighted adversarial imitation learning, providing well-shaped rewards for learning. 

\subsection{Learning a measure of progress by ranking}
\label{sec:rank2learn}
The key assumption that we make in this work is that the provided demonstrations are optimal and that optimal trajectories make monotonic progress towards the goal. This suggests that the true reward for the task (which is unknown) is positive and non-zero for all states, a common occurrence in a huge variety of problems especially goal-reaching tasks ~\cite{hartikainen20ddl, ghosh21gcsl, liu22gcrl}. For most problems, we can find a valid reward function satisfying this assumption. Formally stated:

\begin{assumption} \label{assumption1} True (unknown) reward $r(s, a) > \epsilon$, where $\epsilon > 0$. This suggests that value functions of optimal policies $V^*(s)$ are monotonically increasing. 
\end{assumption}

However, this true reward is unknown, and IRL techniques can recover ill-shaped and hard-to-optimize rewards $\hat{r}(s)$. Instead, we can leverage \cref{assumption1} to directly measure whether states are making \emph{progress} along a trajectory. More specifically, we make the observation that progress along a trajectory can be measured by simply learning a function that can rank different image frames in a trajectory according to their temporal ordering. To do so, we build on recent work in preference modeling ~\cite{christiano17pref, ouyang2022instructgpt} to learn measures of progress by learning how to rank pairs of frames in terms of their natural ordering. Preference modeling methods such as the Bradley-Terry model ~\cite{bradley1952rank} aim to learn a utility function $\hat{u}(s)$ such that the likelihood of ``preferring" a state $s_i^k$ over a different state $s_j^k$ for some expert trajectory $\tau_k$, is given by $p(s_i^k > s_j^k) = \frac{\exp{\hat{u}_\theta(s_i^k)}}{\exp{\hat{u}_\theta(s_j^k)} + \exp{\hat{u}_\theta(s_i^k)}}$. 

Framing temporal ranking as preference modeling allows us to, without any additional human annotation, generate a set of preference labels for trajectories in the expert dataset $\D^e$ given a sampled pairs of states, $(s_i^k, s_j^k)$ along the same expert trajectory. Along $\tau_k$, $s_i^k$ is preferred over $s_j^k$ if it occurs later (i.e., $i > j$). According to the Bradley-Terry model, this suggests that $s_i^k$ should have a higher reward than $s_j^k$ (i.e., $\hat{u}_\theta(s_i^k) > \hat{u}_\theta(s_j^k)$), thereby incentivizing progress along a trajectory. This paradigm naturally lends itself to a training objective for $\hat{u}_\theta$ by simply finding parameters $\theta$ that maximize likelihood over the preference model, resulting in simple cross-entropy classification:  

\vspace{-0.4cm}
\begin{align*}
    \max_{\theta}
        \mathbb{E}_{\substack{\tau_k \sim \D^e \\ s_i^k, s_j^k, \sim \tau_k}} \Biggl[
            \mathds{1}_{i > j} \Bigl[ \log \frac{\exp{\hat{u}_\theta(s_i^k)}}{\exp{\hat{u}_\theta(s_j^k)} + \exp{\hat{u}_\theta(s_i^k)}} \Bigr] + \\ 
            \mathds{1}_{i < j} \Bigl[
            \log \frac{\exp{\hat{u}_\theta(s_j^k)}}{\exp{\hat{u}_\theta(s_j^k)} + \exp{\hat{u}_\theta(s_i^k)}}
            \Bigr]
        \Biggr] \numberthis \label{eq:BCEranking}
\end{align*}

This training objective results in learning a utility function $\hat{u}_\theta$  that is monotonically increasing along a trajectory. This utility function can naturally be converted into a reward function by noting that policy optimization aims to learn policies that maximize the likelihood (and thereby log-likelihood) of progress. By utilizing the likelihood under the Bradley-Terry model and setting the utility of the start of the trajectory $\hat{u}(s_0) = 0$, the likelihood that a state $s$ makes progress over the initial state $s_0$ can be written as $p(s > s_0) = \frac{\exp{\hat{u}_\theta(s)}}{\exp{\hat{u}_\theta(s_0)} + \exp{\hat{u}_\theta(s)}}
    = \frac{1}{1 + \exp{(\hat{u}_\theta(s_0) - \hat{u}_\theta(s))}} = \frac{1}{1 + \exp{(-\hat{u}_\theta(s))}}$. This is essentially just a sigmoid function applied to the learned ranking utilities $\hat{u}(s)$, appropriately normalizing it. For the sake of notation, we denote the likelihood of making progress $p_{\text{RF}}(s) = \frac{1}{1 + \exp{(-\hat{u}_\theta(s))}}$. 
    
A policy that maximizes the log-likelihood of progress of all states can then be obtained as $\max_{\pi} \mathbb{E}_{\pi}\left[\log \cup_{i=1}^n p(s_i > s_0)\right] = \mathbb{E}_{\pi}\left[\sum_{i=1}^n \log p_{\text{RF}}(s_i))\right] = \mathbb{E}_{s, a \sim d^\pi(s, a)}\left[\log p_{\text{RF}}(s) \right]$, where we use the state-action marginal form of the value function. This objective is amenable to any standard policy optimization framework such as model-free reinforcement learning ~\cite{haarnoja18sac, yarats2021image, smith2022walk}, with $\hat{r}(s) = \log p_{\text{RF}}(s) $ as the reward. Note that since it is monotonically increasing along optimal trajectories, $p_{\text{RF}}$ is a value function-like measure rather than a sparse reward type of measure. As discussed in prior work~\cite{hartikainen20ddl}, optimizing value function-like measures can lead to policies that both learn quickly and have bounded suboptimality. The ranking function $\hat{u}_\theta(s)$ can be trained via Eq~\ref{eq:BCEranking} solely from expert trajectories, and the policy can be then optimized with $\hat{r}(s)$. 

\subsection{Incorporating learned rankings into policy optimization}
\label{sec:policyopt}

Naively optimizing this objective naturally leads to the policy exploiting the learned reward function in ``out-of-distribution" states. Since the reward function $\hat{r}(s)$ has only been learned on states from the expert dataset, $s \sim \D^e$, it may overestimate rewards at other states leading to arbitrarily incorrect policies. To remedy this, we incorporate the idea of pessimism into our reward learning framework~\cite{kumar20cql,fujimoto19bcq,offlinerlsurvey} --- penalizing the policy for deviation from the training distribution. Given the expert dataset $\D^e$ and it's corresponding state marginal $d^e(s)$, as well as the current policy $\pi$ and it's state marginal $d^\pi(s)$, we can formulate a pessimistic policy optimization objective as: 

\vspace{-0.4cm}
\begin{equation}
    \label{eq:pessimisticpolopt}
    \max_{\pi} \mathbb{E}_{s \sim d^\pi, a \sim \pi(a|s)}\left[\log p_{\text{RF}}(s) \right] - \alpha D_{KL}(d^\pi(s), d^e(s))
\end{equation}

This objective aims to maximize the likelihood of progress as defined in \cref{sec:rank2learn}. However, it does so while ensuring that the state marginal of the policy remains close to the expert data distribution via a penalty on the divergence $D_{\text{KL}}(d^\pi(s), d^e(s)) = \mathbb{E}_{s \sim d^\pi(s)}\left[ \log \frac{d^\pi(s)}{d^e(s)} \right]$ between the marginal densities of the policy and the expert data. 

This objective is challenging to optimize since the likelihoods under the expert marginal data distribution $d^e(s)$ and the policy marginal distribution $d^\pi(s)$ are both unknown and require density estimation ~\cite{kingma2016iaf, van2016conditional, sohn2015cvae}. Instead, we will show that the objective in \cref{eq:pessimisticpolopt} can be recast as a weighted adversarial imitation learning algorithm that circumvents explicit density estimation.  By substituting the definition of the KL divergence in \cref{eq:pessimisticpolopt} we have the following objective:

\vspace{-0.4cm}
\begin{align*}
    \max_{\pi} \mathbb{E}_{s, a \sim d^\pi}\left[\log p_{\text{RF}}(s) \right] - \alpha \mathbb{E}_{s \sim d^\pi(s)}\left[ \log \frac{d^\pi(s)}{d^e(s)} \right] \\
    = \mathbb{E}_{s, a \sim d^\pi}\left[\log \Bigg(p_{\text{RF}}(s)\bigg(\frac{d^e(s)}{d^\pi(s)}\bigg)^\alpha \Bigg) \right] \numberthis \label{eq:kl-sub}
\end{align*}

To estimate the density ratio $\frac{d^e(s)}{d^\pi(s)}$, we can make use of the fact that despite the likelihoods of $d^e(s)$ and $d^\pi(s)$ not being known, given samples from distributions $d^e(s)$, $d^\pi(s)$, a classifier trained to distinguish between this samples can be used to estimate a density ratio. A classifier $D_\phi(s)$ trained to distinguish between $d^e(s)$, $d^\pi(s)$ can provide us $\frac{d^e(s)}{d^\pi(s)} = \frac{D_\phi(s)}{1 - D_\phi(s)}$ ~\cite{eysenbach21offdynamics}. This reduces to: 
    
\vspace{-0.4cm}
\begin{equation}
\label{eq:finalobj}
    \max_{\pi} \mathbb{E}_{s, a \sim d^\pi}\left[\log \Bigg(p_{\text{RF}}(s) \bigg(\frac{D_\phi(s)}{1 - D_\phi(s)}\bigg)^\alpha \Bigg)\right]
\end{equation}

This equivalence suggests a simple algorithm for optimizing \cref{eq:finalobj} - alternate between (1) training a classifier $D_\phi(s)$ to distinguish between states drawn from the expert video demonstrations and on-policy data collected by the policy $\pi$ using standard binary classification with cross-entropy, and (2) perform policy optimization combining the learned classifier $D_\phi(s)$ together with the learned ranking function in $\hat{r}(s) = p_{\text{RF}}(s) \big(\frac{D_\phi(s)}{1 - D_\phi(s)}\big)^\alpha$. This is similar to adversarial imitation learning methods like GAIL~\cite{ho_generative_2016} but weighted with a learned ranking function $p_{\text{RF}}$. This weighting is crucial, because it provides the policy learning procedure with a dense, shaped reward that is able to guide exploration and encourages efficient, performant policy learning. 

\begin{algorithm}[ht]
\caption{Rank2Reward}
\label{alg:rank2reward}
\begin{algorithmic}[1]
    \STATE \textbf{Require:} Expert demonstration data $\D^e = \{\tau_k\}_{k=1}^N$
    \STATE Initialize policy $\pi$, empty replay buffer $\mathcal{D}_{\text{RB}}$
    \STATE Initialize utility $\hat{u}_\theta$ and classifier $D_\phi$ functions for $\hat{r}(s)$.

    \STATE \textcolor{blue}{// Train the utility ranking function $\hat{u}_\theta$}
    \FOR{step $n$ in \{1, \dots, $N_{\text{ranking}}$\}}
        \STATE Sample state pairs $s_{i}^k, s_{j}^k$ from each trajectory, $\tau_k$
        \STATE Learn $\hat{u}_\theta$ with batch $\{ (s_{t_1}, s_{t_2})_k \}_{k=1}^{bs}$ using \cref{eq:BCEranking}
    \ENDFOR

    \STATE \textcolor{blue}{// Joint policy optimization and reward learning}
    \FOR{step $n$ in \{1, \dots, N\}}
        \STATE With $\pi$, collect transitions $\{\tau_l\}_{l=1}^M$ and store in $\mathcal{D}_{\text{RB}}$
        \IF{$n\, \% \,\texttt{reward\_update\_frequency} == 0$}
            \STATE Sample batch of states $s_e$ from expert $\D^e$
            \STATE Sample batch of states $s_{\pi}$ from replay buffer $\mathcal{D}_{\text{RB}}$ 
            \STATE Update $D_\phi$ to classify $s_e$ from $s_\pi$ with BCE. 
        \ENDIF
        \STATE Sample batch of transitions $s_\pi$ from $\mathcal{D}_{\text{RB}}$
        \STATE Update $\pi$ to maximize returns using \cref{eq:final-reward}
    \ENDFOR
\end{algorithmic}
\end{algorithm}

\begin{figure}[!h]
    \centering
    \includegraphics[width=0.8\linewidth]{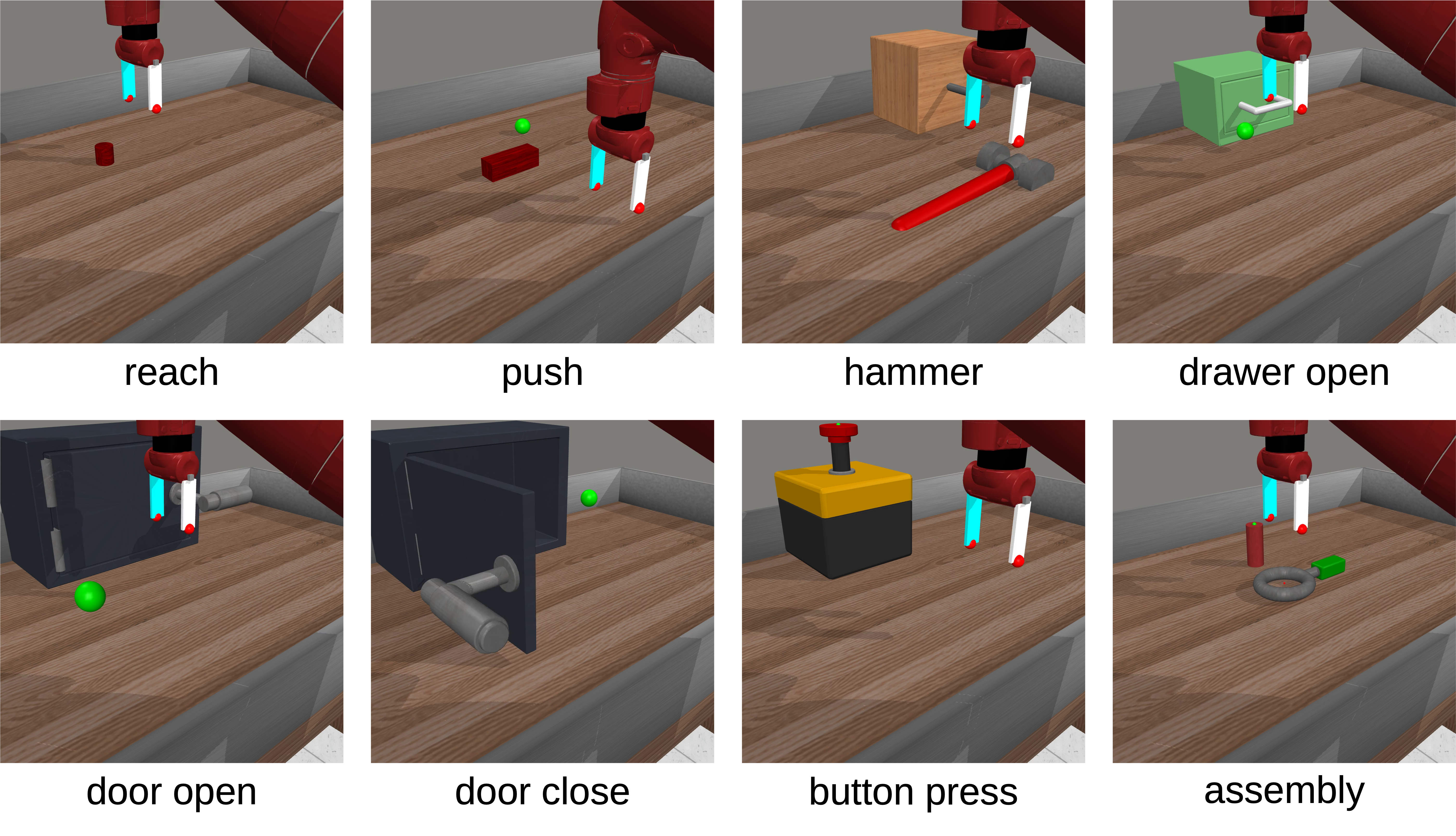}
    \caption{\footnotesize{Simulation environments for evaluation in the Meta-world~\cite{yu2020meta} benchmark: (1) reach, (2) push, (3) hammer, (4) drawer open, (5) door open, (6) door close, (7) button press, (8) assembly}}
    \label{fig:metaworld_tasks}
    \vspace{-0.4cm}
\end{figure}

\begin{figure}[!t]
    \centering
    \includegraphics[width=\linewidth]{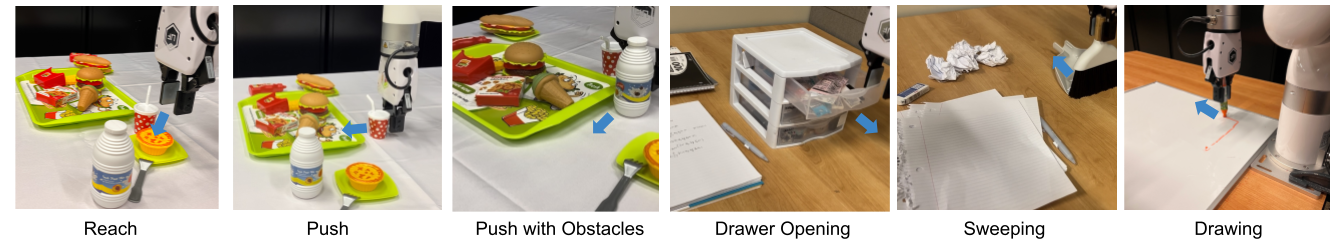}
    \caption{\footnotesize{Real-world environments including standard tasks (reaching and pushing), tasks where exploration is non-trivial (pushing with obstacles and drawer opening), and tasks where state estimation is non-trivial (sweeping and drawing). The blue arrows indicate the directions to go.}}
    \label{fig:realenvs}
    \vspace{-0.6cm}
\end{figure}

\subsection{Practical algorithm overview}
In \Cref{alg:rank2reward}, we show how our method for learning a reward function can be used with any off the shelf reinforcement learning algorithm - here, off-policy learning methods \cite{yarats2021image, smith2022walk} for data efficiency. Notably, the ranking component $p_{\text{RF}}(s)$ of the reward can be learned offline solely from expert data, independent of learning the policy, and only the classification component $D_{\theta}(s)$ depends on both the expert data and data collected by the current learned policy. Our final simplified learned, estimated reward function is: 

\vspace{-0.4cm}
\begin{equation}
\label{eq:final-reward}
    \hat{r}(s) = \log p_{\text{RF}}(s) + \alpha \Bigl( \log D_\phi(s) - \log (1 - D_\phi(s) ) \Bigr)
\end{equation}

%% file: sections/results.tex
\begin{table*}[!h]
    \begin{center}
    \begin{tabular}{|| p{2.8 cm} p{1.5 cm} p{1.3 cm} p{1.3 cm} p{1.3 cm} p{1.6 cm} p{1.3 cm} ||}
     \hline
      Task & Reaching & Pushing & Pushing w/ Obst & Drawer Opening & Sweeping & Drawing \\
     \hline
     Metric & $L2$-dist & $L2$-dist & StS & StS & Incompletion & StS \\
     \hline\hline
     \Method (Ours) & \textbf{0.31} & \textbf{0} & \textbf{4391} & \textbf{6079} & \textbf{0\%} & \textbf{79} \\
     GAIL & 0.43 & 1.26 & FAILED & FAILED & 60\% & FAILED \\
     \hline
    \end{tabular}
    \end{center}
     \caption{\footnotesize{Evaluation results of real-world training using \Method, with a comparison to GAIL~\cite{ho_generative_2016}. For all metrics below, lower is better. $L2$-dist is the $L2$-distance in centimeters (cm) from the goal state at the end. Steps to Success (StS) measures the number of environment steps required to successfully learn the task. Incompletion refers to the percentage of objects that were not successfully swept off the table. We see that \Method is able to consistently outperform objectives that do not provide shaped reward.}}
     \label{table:table1}
\end{table*}

\section{Experimental Results}
\label{sec:result}

In our experiments, we evaluate our proposed technique for learning reward functions from video demonstrations on simulated and real-world manipulation tasks. We also experiment with scaling our approach to internet-scale in-the-wild data from Ego4D \cite{Ego4D2022CVPR}. In simulation, we leverage a model-free RL method for learning from images, DrQv2 \cite{yarats2021drqv2, yarats2021image}, while in the real world, we leverage a data-efficient actor-critic technique \cite{smith2022walk, hiraoka2021dropout}. All experiments use a frozen pre-trained visual feature extractor from \cite{nair_r3m_2022}.

\subsection{Evaluation Environments}
\label{sec:evalenvs}
We evaluate \Method on the following environments: 

\textbf{Meta-world \cite{yu2020meta}:} This simulation benchmark consists of a table-top Sawyer robot arm, as shown in \cref{fig:metaworld_tasks}. We instantiate environments with random initial states and use image-based observations. We compare our method against baselines by measuring the average episodic return based on the hand-defined rewards available with the benchmark.

\textbf{Real-World xArm Environment:}
We utilize a tabletop mounted 5 DoF xArm5 manipulator. We perform end-effector positional control, where the action space is normalized delta positions, and use image-based observations. We test \Method on 6 real-world tasks, as shown in \cref{fig:realenvs}. Our more complex tasks highlight situations where exploration is non-trivial, techniques like object tracking are ineffective, and reward specification overall is difficult. 

\subsection{Baseline Comparisons}
We compare \Method with the following baselines: \textbf{(1) GAIL}~\cite{ho_generative_2016} Reward function is a classifier of whether state comes from expert demonstration trajectories. This is similar to our method \textit{without} the ranking term and is akin to ensuring the policy state visitation distribution matches that of the expert data. \textbf{(2) AIRL}~\cite{fu2017learning} The reward function is similar to GAIL above but scaled with $r_{\text{AIRL}} = \log(r_{\text{GAIL}}) - \log(1-r_{\text{GAIL}})$. \textbf{(3) VICE}~\cite{fu_variational_2018} The reward function is similar to GAIL above but instead of learning a classifier of whether the state comes from the expert demonstration \textit{trajectories}, VICE classifies whether a state is the expert \textit{goal state}. \textbf{(4) SOIL}~\cite{radosavovic_state-only_2021} Learn an inverse dynamics model and uses the data to infer actions for the expert states and uses this for imitation learning. \textbf{(5) TCN}~\cite{sermanet2018time} We utilize the single-view variant of TCN to learn an embedding space and perform feature tracking with an expert demonstration to generate rewards. \textbf{(6) ROT}~\cite{haldar2022watch} A recent method using on optimal transport-based trajectory matching. To compare in a similar setting, we do not utilize the behavioral cloning initialization and regularization components of ROT, as our method presumes no access to expert actions. \textbf{(7) Ranking only} This ablation uses only the ranking function, without the adversarial classifier, to generate reward.

\begin{figure*}[!t]
    \centering
    \includegraphics[width=\linewidth]{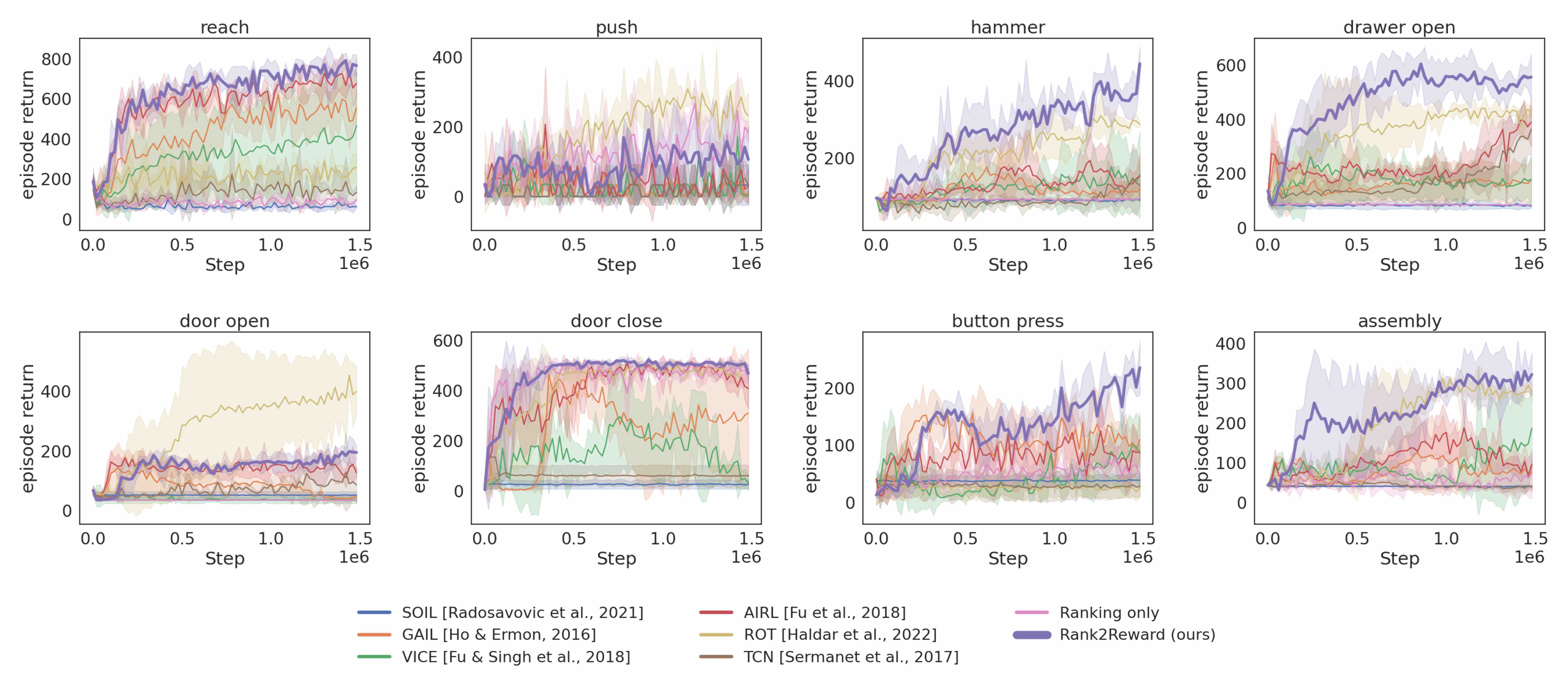}
    \caption{\footnotesize{Visualization of policy learning experiments in simulation with \Method. Our method - \Method (purple) distinctly outperforms other methods on \texttt{hammer}, \texttt{drawer open}, \texttt{button press}, and \texttt{assembly}, while performing similarly to the best baseline with \texttt{reach}, and \texttt{door close} and worse than the best baseline in \texttt{push} and \texttt{door open}. The plots show the episodic return from 10 evaluation episodes averaged over 3 seeds plotted over the course of training the DrQ-v2 agent for 1.5 million steps, with higher being better.}}
    \label{fig:metaworld_returns}
\end{figure*}

\subsection{Simulated Experiments}
\label{sec:simulated_exps}

To quantify performance, we examine the average episodic return from 10 evaluation episodes averaged over 3 seeds as shown in \cref{fig:metaworld_returns}. While all methods achieve non-zero returns, our method learns quickly and more effectively than baselines for state-only imitation learning and reward assignment. Performance on \texttt{hammer}, \texttt{drawer open}, \texttt{button press}, and \texttt{assembly} is significantly better than baselines, while learning curves on \texttt{reach}, \texttt{push}, \texttt{door open} show comparable or slightly better performance between our method and AIRL~\cite{fu2017learning}. Notably, TCN~\cite{sermanet2018time} and ROT~\cite{haldar2022watch} are methods designed for learning from video whereas most other methods focus on low dimensional states which makes their comparisons more insightful. Our method performs similarly or slightly better than ROT in most environments. However, in \texttt{door open} and \texttt{push}, ROT noticeably outperforms our method, while in the simplest domain \texttt{reach}, ROT struggles to achieve high rewards. In all domains TCN performs similarly to some baseline methods, but does not achieve comparable performance to \Method.

\subsection{Real-World Robotic Experiments}

We evaluate \Method on real-world tasks (\cref{sec:evalenvs}) and compare our performance with GAIL~\cite{ho_generative_2016} as shown in \Cref{table:table1}. We use different success metrics for our evaluation tasks --- distance to goal in reaching and pushing, number of environment steps required to learn the task for pushing with obstacle, drawer opening, and drawing, and percent of objects not successfully swept for sweeping. \Method is able to effectively learn behaviors across real-world robotics domains purely from image observations and video demonstrations. Our baselines are unable to reliably learn any of our more complex tasks beyond reaching and pushing, while \Method can learn in under 2 hours of real-world interaction even for the more challenging tasks. 

\subsection{Ego4D experiments}

\begin{figure*}[!t]
    \centering
    \includegraphics[width=\linewidth]{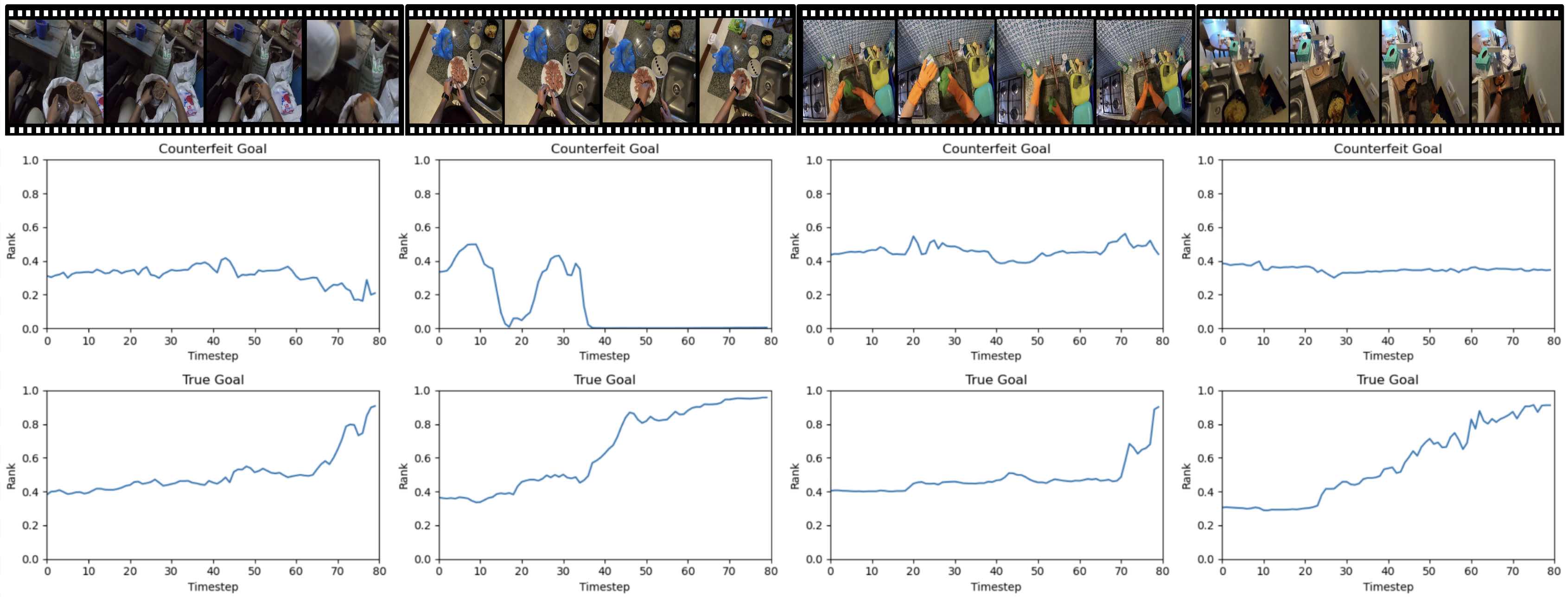}
    \caption{\footnotesize{Visualization of \Method output when four random videos from Ego4D are evaluated for the true goal and a counterfactual goal that does not correspond to the input.\textbf{(Top):} Reward with a counterfeit goal. The reward does not consistently increase in this setting, as expected. \textbf{(Bottom):} Reward for a trajectory with a true goal image. The reward in this case is monotonically increasing, demonstrating the ability to extract well-shaped, dense rewards.}}
    \label{fig:ego4d_res}
    \vspace{-0.5cm}
\end{figure*}

To show that \Method is a generalizable and scalable paradigm, we apply our approach to the Hand \& Object Interactions data from Ego4D \cite{Ego4D2022CVPR}. From 27,000 segments processed at 10 fps, we have 2.16 million frames. We train with 20,000 segments, and leave the rest for evaluation. From each clip, we utilize the last frame as the goal frame and learn a ranking component conditioned on the goal frame. 

For the discriminator, we sample a positive frame from the same clip as the goal and a negative frame from a different clip as the goal, and train $D_\phi$ to classify whether a given frame and the goal frame come from the same video. These negative frames with goals that do not match can be thought of as counterfactual examples. We randomly select four segments from the evaluation set and present the output of \Method when evaluated with the true goal and a counterfactual goal in \cref{fig:ego4d_res}. When a state is evaluated with the true goal, reward is overall increasing whereas when evaluated with a counterfactual goal, it is both not increasing and has an overall lower value. Such a defined and well-shaped reward landscape on diverse, real-world data holds promising value in lowering the difficulty of providing expert data to learn robotic tasks.

\begin{figure}[!h]
    \centering    \includegraphics[width=\linewidth]{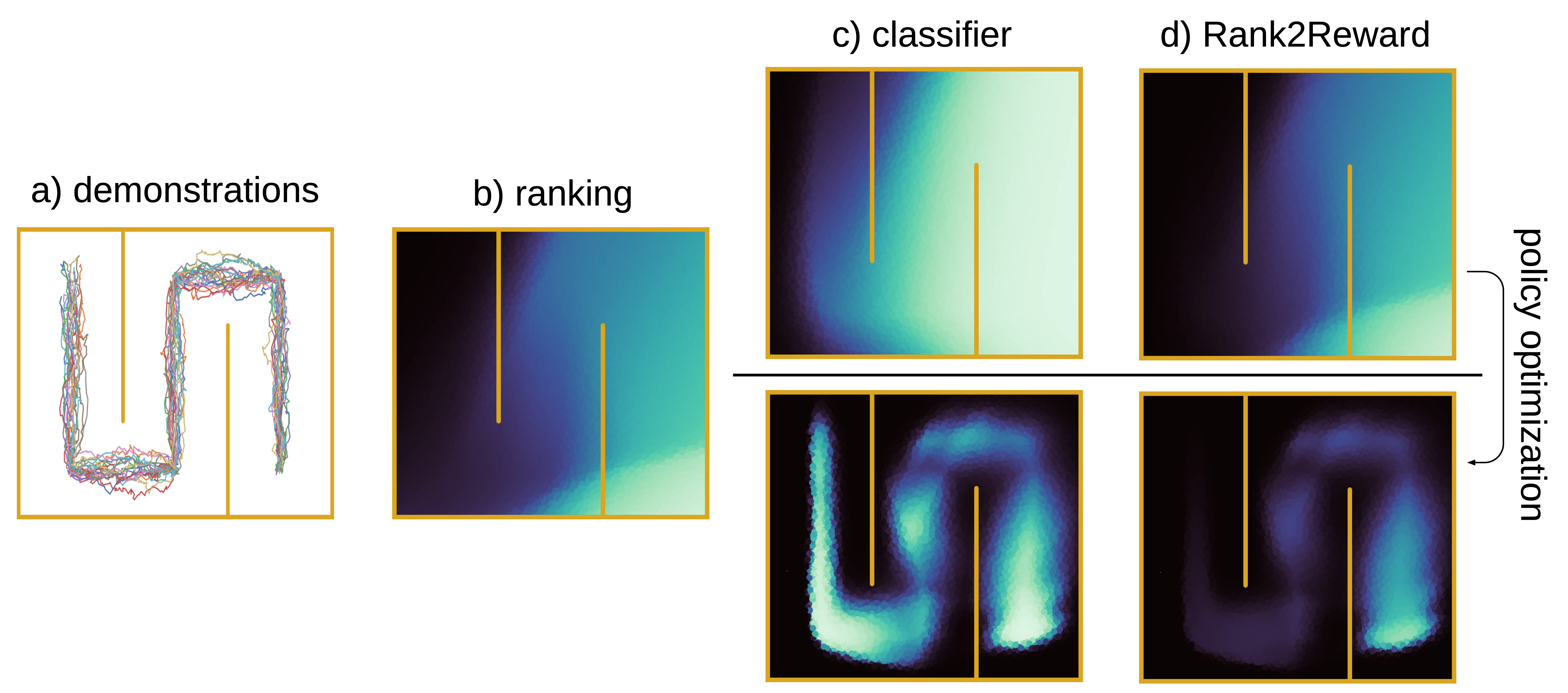}
    \caption{\footnotesize{Visualization of different components of \Method in a two-wall 2D maze environment where state is position. (a) Demonstration data (b) $p_{\text{RF}}(s)$. Note the spurious high values assigned to regions not visited by the expert. (c) $D_\phi(s)$, before and after policy optimization. Note the induced down weighting of out of expert distribution data (d) \Method, $D_\phi(s) * p_{\text{RF}}(s)$, before and after policy optimization.}}
    \label{fig:two-wall-maze}
    \vspace{-0.4cm}
\end{figure}

\subsection{Reward function analysis}
\label{sec:reward_viz}

We visualize the shaping of our reward function over policy optimization in a continuous 2D two-wall maze environment, where the agent starts at the top left and the goal is at the bottom right. Given 20 expert demonstrations (\cref{fig:two-wall-maze}a), we visualize the fixed ranking function over the landscape where greedily moving towards highly ranked states does not necessarily lead to the goal (\cref{fig:two-wall-maze}b). We show the evolution of the classifier during policy optimization (\cref{fig:two-wall-maze}c, top and bottom). Initially, the classifier gives higher values to states on the right half of the maze with some slight shaping from states likely to be stumbled upon by random exploration. However, over RL the classifier better distinguishes states similar to the expert demonstrations from other states. When combined with our ranking function which was well-shaped but had spuriously high output for out of distribution inputs, we see that our final reward function is both well-shaped and well-defined across the whole state space (\cref{fig:two-wall-maze}d, bottom).

%% file: sections/discussion.tex
\section{Conclusion and Limitations}
\label{sec:conclusion}

In this work, we have shown that learning how to rank visual observations from a demonstration can be used to infer well-shaped reward functions when paired with ideas from pessimism. \Method is simple to use, easily integrating into many popular off the shelf RL algorithms. By combining how to rank with how to classify expert demonstration data from policy-collected data, our learned reward function is interpretable yet performant. We show experimental results in both simulated and real-world robotic domains showing the efficacy of this technique in robotic manipulation settings. The key benefits of a reward inference technique like \Method really lies in it's simplicity and ease of use. 

\textbf{Limitations and Future Work} 
There are several limitations with \Method that naturally lead to future work. Notably, there is an embodiment shift between human demonstration videos like those found in \cite{Ego4D2022CVPR, Damen2018EPICKITCHENS} and our robot manipulators. To make use of internet-scale data, we must use representations that generalize across manipulators, perhaps building on ~\cite{zakka_xirl_2021, rong2021frankmocap, alakuijala2023learning}. Secondly, the rewards trained here are still single-task and it would be challenging to have a different reward and agent for every task. Thirdly, as of right now the classifier $D_\phi$ is sensitive to changes in the background and dynamic scenes. Incorporating pre-trained visual representations into this process can be very effective at ensuring generalizable, robust reward inference. Real world deployment will require further invariant representations during reward inference. And lastly, there are challenges with stabilizing and using adversarial optimization in a stable way. Future work can build on advances in contrastive learning and large scale GAN-training to make this process more scalable. 